\documentclass{article}
\usepackage{ijcai26}
\usepackage{times}
\usepackage{soul}
\usepackage{url}
\usepackage[hidelinks]{hyperref}
\usepackage[utf8]{inputenc}
\usepackage[small]{caption}
\usepackage{graphicx}
\usepackage{subfigure} 
\usepackage{amsmath}
\usepackage{amsthm}
\usepackage{amssymb}
\usepackage{booktabs}
\usepackage{algorithm}
\usepackage{algorithmic}
\usepackage{float}
\usepackage{listings}
\usepackage[switch]{lineno}
\usepackage{tcolorbox}
\tcbuselibrary{breakable, skins}

\newtcolorbox[auto counter, number within=section]{purposebox}[2][]{%
  colback=white, 
  colframe=blue!50!black, 
  width=\textwidth,
  arc=2mm, 
  boxrule=0.5mm, 
  title={\normalsize #2},
  breakable, 
  fonttitle=\bfseries\Large, 
  fontupper=\small, 
  #1
}

\title{Beyond Rule-Based Workflows: An Information-Flow-Orchestrated Multi-Agents Paradigm via Agent-to-Agent Communication from CORAL}

\author{%
  \textbf{Xinxing Ren}$^{1,2,*}$ \quad
  \textbf{Quagmire Zang}$^{3,*}$ \quad
  \textbf{Caelum Forder}$^{1,*}$ \quad
  \textbf{Suman Deb}$^{1,*}$ \quad
  \textbf{Ahsen Tahir}$^{1,5,*}$ \\
  \textbf{Roman J. Georgio}$^{1}$ \quad
  \textbf{Peter Carroll}$^{1}$ \quad
  \textbf{Zekun Guo}$^{4,\dagger}$ \\[6pt]
  $^{1}$ Coral Protocol \quad
  $^{2}$ Brunel University of London \quad
  $^{3}$ Universitéit Lëtzebuerg \\
  $^{4}$ University of Hull \quad
  $^{5}$ National University of Computer and Emerging Sciences \\[6pt]
  $^{*}$ Equal contribution \quad
  $^{\dagger}$ Corresponding author
}

\begin{document}

\maketitle

\begin{abstract}
    Most existing Large Language Model (LLM)–based Multi-Agent Systems (MAS) rely on predefined workflows, where human engineers enumerate task states in advance and specify routing rules and contextual injections accordingly. Such workflow-driven designs are essentially rule-based decision trees, which suffer from two fundamental limitations: they require substantial manual effort to anticipate and encode possible task states, and they cannot exhaustively cover the state space of complex real-world tasks. To address these issues, we propose an Information-Flow-Orchestrated Multi-Agent Paradigm via Agent-to-Agent (A2A) Communication from CORAL, in which a dedicated information flow orchestrator continuously monitors task progress and dynamically coordinates other agents through the A2A toolkit using natural language, without relying on predefined workflows. We evaluate our approach on the general-purpose benchmark GAIA, using the representative workflow-based MAS OWL as the baseline while controlling for agent roles and underlying models. Under the pass@1 setting, our method achieves 63.64\% accuracy, outperforming OWL’s 55.15\% by 8.49 percentage points with comparable token consumption. Further case-level analysis shows that our paradigm enables more flexible task monitoring and more robust handling of edge cases. Our implementation is publicly available at: \url{https://github.com/Coral-Protocol/Beyond-Rule-Based-Workflows}.
\end{abstract}

\section{Introduction}

Recent advances in LLMs have enabled the development of intelligent agents capable of performing complex reasoning and decision-making tasks. Following the Agentic Benchmark Checklist proposed in \cite{zhu2025establishing}, agentic tasks are characterized by: (i) sustained multi-step interactions with an external environment, (ii) iterative information gathering under partial observability, and (iii) adaptive strategy refinement based on environmental feedback. Such agentic systems have achieved remarkable performance in a wide range of applications, including code generation \cite{hong2023metagpt,yang2024swe}, web browsing \cite{wei2025browsecomp}, finance \cite{yu2025finmem}, and scientific discovery \cite{su-etal-2025-many}. As task complexity increases, many agentic problems naturally require diverse expertise and coordinated decision making, which has motivated a growing research focus on llm-based MAS \cite{ijcai2024p890}. Collaborative multi-agent approaches such as OWL \cite{hu2025owl} and MetaGPT \cite{hong2023metagpt} have demonstrated that coordinated MAS can outperform single-agent systems on complex, general-purpose tasks requiring heterogeneous skill sets as well as on challenging code generation problems. 

Most existing MAS are constructed using predefined workflows, as exemplified by OWL \cite{hu2025owl}, MetaGPT \cite{hong2023metagpt}, and AutoAgent \cite{tang2025autoagent}. 
Fundamentally, a workflow can be viewed as a rule-based decision tree, where human engineers predefine discrete task states and specify routing policies and contextual injections conditioned on these states. 
While effective for well-scoped tasks, this paradigm suffers from inherent limitations: 
(1) it requires substantial manual effort to anticipate task states and design corresponding routing logic; and 
(2) for complex real-world tasks, it is theoretically infeasible to exhaustively enumerate all possible states in advance.
As a result, workflow-based supervision often struggles to reliably monitor task execution and handle unforeseen edge cases.

\begin{figure}[t]
    \centering
    \includegraphics[width=\linewidth]{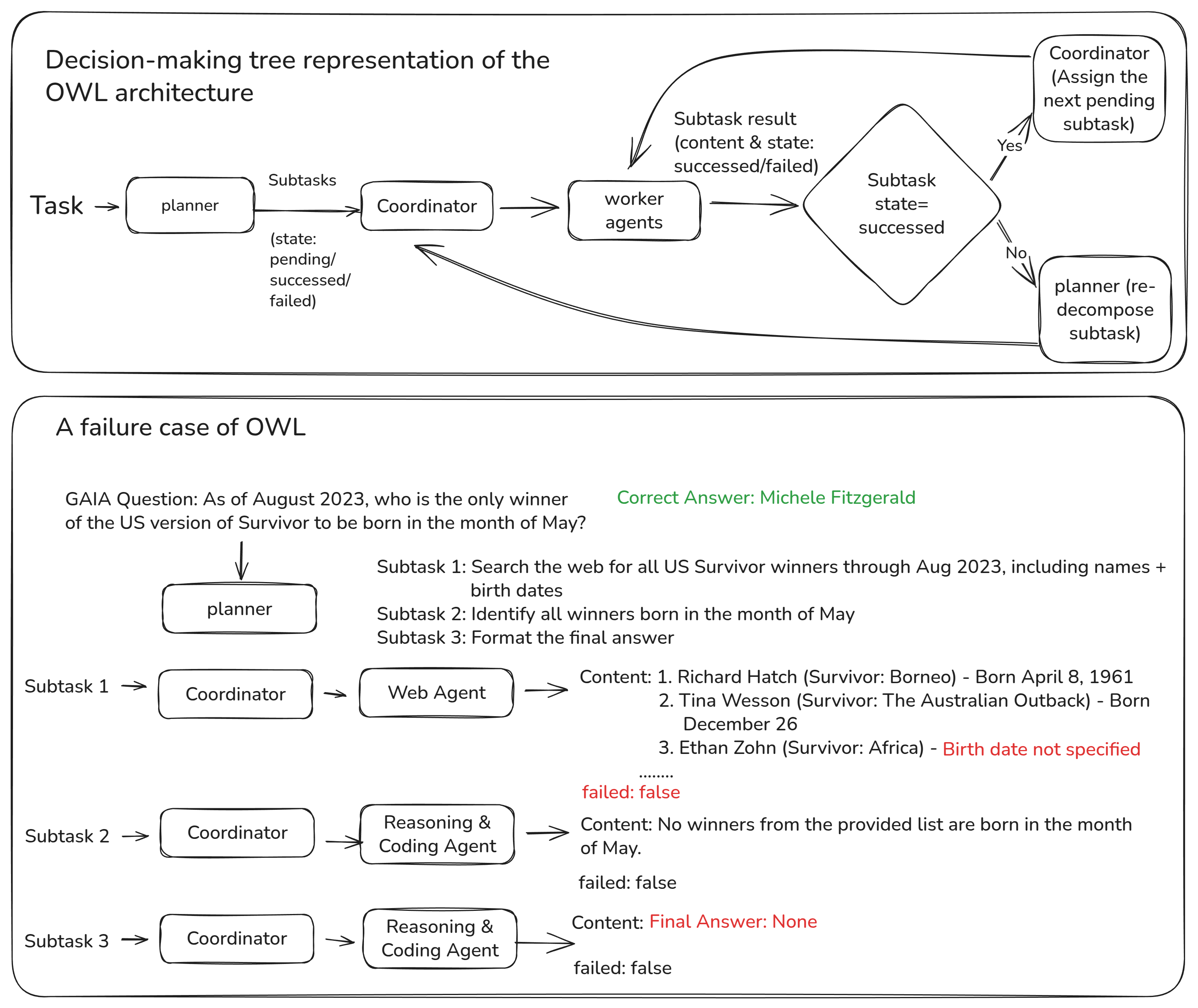}
    \caption{Decision-making tree representation of the OWL architecture.}
    \label{fig:owl_structure}
\end{figure}

Figure~\ref{fig:owl_structure} illustrates the representative workflow-based MAS design of OWL. The upper part shows a decision-making tree representation, where tasks are decomposed into stateful subtasks and iteratively routed to worker agents, with the entire task re-decomposed upon subtask failure. While this design provides a clear control flow, it relies on a predefined set of task states and routing rules that must be manually specified by human engineers. The lower part presents a representative failure case: a web agent retrieves all U.S. Survivor winners but fails to obtain birth dates for some of them while still marking the subtask as successful, causing subsequent subtasks to proceed on incomplete information and leading to an incorrect final answer. This partial fulfillment makes it difficult for the agent to determine whether the subtask should be considered a success or a failure, highlighting a core limitation of workflow-based supervision, that a limited set of predefined states is insufficient to monitor the full task execution process, while exhaustively anticipating and encoding edge cases remains infeasible for human engineers.

\begin{figure*}[t]
    \centering
    \includegraphics[width=\textwidth]{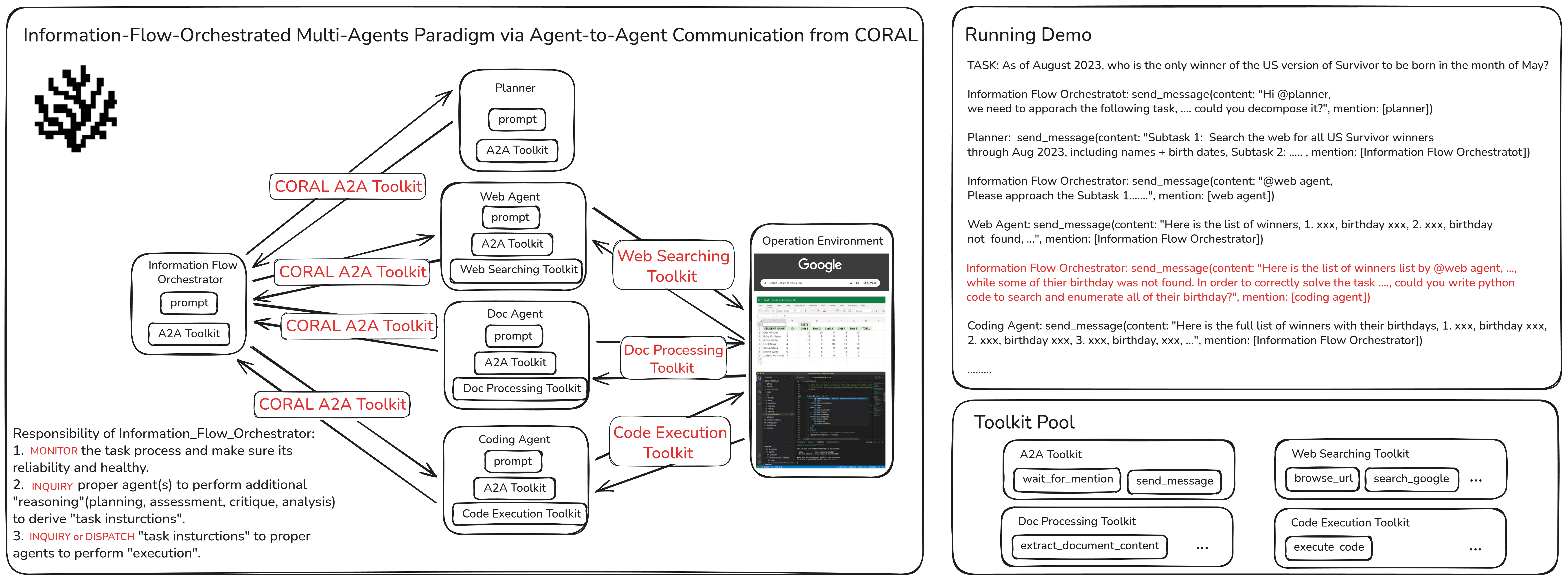}
    \caption{Overview of the proposed Information-Flow-Orchestrated Multi-Agent Paradigm via Agent-to-Agent (A2A) Communication. A dedicated information flow orchestrator monitors task progress and dynamically coordinates agents through natural-language A2A interactions, eliminating the need for predefined workflows.}
    \label{fig:coral_architecture}
\end{figure*}

Motivated by these observations, \textit{we ask whether it is possible to transfer the responsibility of constructing and supervising workflows from human engineers to the agents themselves.} A similar shift has occurred in autonomous driving \cite{chen2024end}, moving from handcrafted state machines to end-to-end perception-driven control \cite{prakash2021multi,hu2023planning}, motivating a corresponding transition toward agent-driven coordination in MAS. To realize this vision, we propose an \textbf{Information-Flow-Orchestrated Multi-Agent Paradigm via A2A Communication Toolkit}. As illustrated in Figure~\ref{fig:coral_architecture}, a dedicated information flow orchestrator continuously monitors task progress and dynamically coordinates other agents through the A2A communication toolkit from CORAL using natural language, without relying on predefined workflows.

To validate the effectiveness of our proposed paradigm, we evaluate it on the general-purpose benchmark GAIA \cite{mialon2024gaia}, using the classical workflow-based multi-agent system OWL as the baseline. For a fair comparison, both systems employ identical agent roles and the same underlying large language models. Under the pass@1 evaluation setting, our method achieves an accuracy of 63.64\%, outperforming OWL’s 55.15\% by 8.49 percentage points with comparable token consumption. Further case-level analysis shows that our paradigm enables more flexible task monitoring and more robust handling of edge cases.

Our contributions can be summarized as follows:
\begin{itemize}
    \item We propose an \textbf{Information-Flow-Orchestrated Multi-Agent Paradigm via A2A Communication}, which shifts workflow construction and supervision from human-designed state machines to agent-driven supervision and coordination.
    \item Under a controlled experimental setting where both systems employ identical agent roles and the same underlying language models, our approach achieves \textbf{63.64\%} accuracy on the GAIA benchmark (pass@1), outperforming the classical workflow-based MAS OWL (\textbf{55.15\%}) by \textbf{8.49 percentage points} with comparable token consumption.
    \item Through detailed case-level analysis, we demonstrate that our paradigm enables more flexible task monitoring and exhibits greater robustness to edge cases that are difficult to handle under workflow-based MAS.
\end{itemize}

\section{Related Work and Preliminaries}

\textbf{Multi-Agent Systems.}
Existing MAS can be broadly divided into \emph{domain-specific} and \emph{general-purpose} categories. Domain-specific MAS tailor agent collaboration to particular tasks, such as software engineering (e.g., MetaGPT \cite{hong2023metagpt}, SWE-Search \cite{antoniadesswe}), data analysis (AutoKaggle \cite{li2024autokaggle}), engineering simulation (SimuGen \cite{ren2025simugen}), and scientific idea generation (VIRSCI \cite{su-etal-2025-many}), leveraging strong domain priors and specialized workflows. In contrast, general-purpose MAS, also referred to as \emph{generalist agents} and formalized by the GAIA benchmark \cite{mialon2024gaia}, aim to solve open-ended tasks across domains. Representative systems include OpenAI’s Deep Research, OWL \cite{hu2025owl}, and no-code platforms such as AutoAgent \cite{tang2025autoagent}. Due to their higher variability and unpredictability, general-purpose tasks make it difficult to predefine exhaustive task states, motivating our focus on this setting.

\textbf{Dynamic Multi-Agent Systems.}
As shown in Table~\ref{tab:dynamic_mas_comparison}, recent works have explored the design of topological scaffolds to coordinate groups of agents. GTPSwarm~\cite{zhuge2024gptswarm} formulates agent collaboration as a learnable graph structure, MasRouter~\cite{yue-etal-2025-masrouter} learns embedding spaces to map queries to agents and interaction topologies, and Conductor~\cite{anonymous2025learning} trains a dedicated conductor to jointly perform task decomposition and agent routing. However, these approaches typically determine the MAS topology and routing policies prior to task execution, which limits their ability to monitor emergent edge cases and adapt dynamically during task progression. Puppeteer~\cite{dang2025multiagent} performs step-wise agent routing at runtime. Nevertheless, due to the lack of explicit natural-language instructions to routed agents—relying instead on concatenating previous agents’ outputs into the next agent’s context—its robustness remains limited. In contrast, our proposed \emph{A2A-based paradigm} differs fundamentally from prior work. At each step of task execution, the information flow orchestrator actively monitors task progress and issues explicit, step-specific inquiries or instructions to subsequent agents via the A2A communication toolkit, enabling fine-grained coordination and adaptive handling of emergent edge cases.

\begin{table*}[t]
\centering
\small
\begin{tabular}{lccc}
\toprule
\textbf{Method} 
& \textbf{Dynamic Orchestration} 
& \textbf{Adaptive Routing} 
& \textbf{Explicit Natural Language Instructions} \\
\midrule
GTPSwarm~\cite{zhuge2024gptswarm} 
& $\checkmark$ 
& $\times$ 
& $\times$ \\

MasRouter~\cite{yue-etal-2025-masrouter} 
& $\checkmark$ 
& $\times$ 
& $\times$ \\

Conductor~\cite{anonymous2025learning} 
& $\checkmark$ 
& $\times$ 
& $\times$ \\

Puppeteer~\cite{dang2025multiagent} 
& $\checkmark$ 
& $\checkmark$ 
& $\times$ \\

\textbf{Ours (A2A-based)} 
& $\checkmark$ 
& $\checkmark$ 
& $\checkmark$ \\
\bottomrule
\end{tabular}
\caption{Comparison of dynamic multi-agent systems.}
\label{tab:dynamic_mas_comparison}
\end{table*}

\section{Information-Flow-Centric Multi-Agent Coordination}

\textbf{Agent-to-Agent Communication Toolkit.}
We utilize a set of A2A communication toolkits from CORAL
\begin{equation}
\mathcal{K}^{\text{A2A}} = \{\texttt{wait\_for\_mention}, \texttt{send\_messages}\},
\end{equation}
which are available to all agents for sending and receiving natural-language messages.

The toolkit \texttt{wait\_for\_mention} induces a blocking operation
\begin{equation}
\texttt{wait\_for\_mention}(a_i) \rightarrow m,
\end{equation}
where agent $a_i \in \mathcal{A}$ enters a waiting state until it receives a message $m$ from another agent.

The toolkit \texttt{send\_messages} defines a message-sending operation
\begin{equation}
\texttt{send\_messages}(a_i, a_j, c),
\end{equation}
where agent $a_i$ sends a natural-language message with content $c \in \mathcal{M}$ to a designated agent $a_j \in \mathcal{A}$.

Together, these toolkits induce an asynchronous communication process
\begin{equation}
m_t = (a_i, a_j, c_t),
\end{equation}
enabling agents to actively coordinate through natural-language communication, without relying on human-engineered routing rules or manual context engineering.

\textbf{Information-Flow-Orchestrated Multi-Agent Paradigm.}
We consider a multi-agent system defined by a finite set of agents
\begin{equation}
\mathcal{A} = \{a_1, a_2, \dots, a_N\},
\end{equation}
among which a distinguished agent
\begin{equation}
a_o \in \mathcal{A}
\end{equation}
is designated as the information flow orchestrator.

We impose an asymmetric communication constraint
\begin{equation}
(a_i \rightarrow a_j) \in \mathcal{C} \;\; \Rightarrow \;\; (i = o) \lor (j = o),
\end{equation}
such that the information flow orchestrator may communicate with any agent, while all other agents communicate exclusively with the information flow orchestrator.

The query
\begin{equation}
q \in \mathcal{Q}
\end{equation}
is first received by the information flow orchestrator $a_o$ at time step $t=0$.

At each step $t$, the information flow orchestrator generates a coordination message based on the task query and the accumulated inter-agent communication history. Let
\begin{equation}
\mathcal{H}
\end{equation}
denote the message history between the information flow orchestrator and other agents. The information flow orchestrator’s outgoing message is generated as
\begin{equation}
m_{o,t} \leftarrow f_o(\mathcal{H}, q, p_o),
\end{equation}
where $p_o$ denotes the prompt that specifies the role and responsibilities of the information flow orchestrator, including: (i) monitoring the task execution process to ensure reliability and consistency; (ii) inquiring appropriate agents when additional reasoning is required to derive or refine task instructions; and (iii) relaying or dispatching task instructions to appropriate agents for execution.

The generated message is then sent to a selected agent $a_j \in \mathcal{A}$ in the form
\begin{equation}
m_{o,t}^{o \rightarrow j} = (a_o, a_j, c_{o,t}),
\end{equation}
where the content $c_{o,t}$ is expressed in natural language and takes the form of either an \emph{inquiry} or an \emph{instruction}. The message $m_{o,t}^{o \rightarrow j}$ is appended to the message history $\mathcal{H}$.

Upon receiving a message from the information flow orchestrator, agent $a_j$ either produces a direct response or invokes external tools to obtain intermediate results before responding. Let
\begin{equation}
\tilde{z}_{j,t}
\end{equation}
denote the (optional) result obtained via tool invocation. The agent response is then generated as
\begin{equation}
m_{j,t} \leftarrow f_j(\tilde{z}_{j,t}, \mathcal{H}, p_j),
\end{equation}
where $p_j$ denotes the prompt associated with agent $a_j$, which may vary across agents depending on their roles.

The agent response is sent back to the information flow orchestrator as
\begin{equation}
m_{j,t}^{j \rightarrow o} = (a_j, a_o, c_{j,t}),
\end{equation}
and appended to the message history $\mathcal{H}$.

This interaction process proceeds through iterative message exchanges between the information flow orchestrator and other agents. 

The information flow orchestrator is equipped with a dedicated \texttt{submit\_answer\_tool}, and may decide to submit a final answer based on the accumulated message history $\mathcal{H}$, the original query $q$, and its prompt $p_o$, i.e.,
\begin{equation}
\texttt{submit} \leftarrow f_o(\mathcal{H}, q, p_o).
\end{equation}
The submission criteria is explicitly defined in the information flow orchestrator prompt $p_o$. In addition, a fixed execution-time budget of 30 minutes is enforced, after which the information flow orchestrator is required to submit its current best answer.

\section{Evaluation and Analysis}

\subsection{Benchmark: GAIA}

The more general a task is, the harder it becomes for human engineers to exhaustively anticipate and encode all possible edge cases that may arise during execution. To evaluate our paradigm under such settings, we adopt GAIA~\cite{liu2025joyagent} as the benchmark in this work.

GAIA is a benchmark designed for generalist AI assistants, covering diverse domains and requiring multimodal reasoning, code execution, and live web search. We conduct our evaluation on the GAIA validation set, which consists of 165 tasks with difficulty levels ranging from Level~1 to Level~3. Each task has a unique, objectively verifiable ground-truth answer.

\subsection{Baseline and Experimental Settings}

We formulate the objective of our experiments around two research questions (RQs):

\textbf{RQ1}: Can our A2A-based MAS paradigm match the performance of a workflow-based MAS in terms of task completion rate and cost?

\textbf{RQ2}: Can our A2A-based MAS paradigm surpass the performance of a workflow-based MAS in terms of task completion rate and cost?

To answer these questions, we choose OWL~\cite{hu2025owl} as the baseline. OWL is a mature and representative workflow-based MAS designed for general-purpose tasks, and it represents the state of the art among open-source MAS evaluated on the GAIA benchmark.

For a fair comparison, we adopt the same set of agent roles as in OWL, including a planner, web agent, document agent, and reasoning \& coding agent, while excluding the coordinator role, as its routing functionality after task decomposition partially overlaps with that of our information flow orchestrator. In our paradigm, these agents are organized under an information flow orchestrator. All agents are equipped with the proposed A2A communication toolkit and corresponding prompts. Detailed descriptions of the toolkit and prompts are provided in Appendix~A.

To evaluate \textbf{RQ1}, we use a strong language model, Grok~4.1~Fast, and assign it to all agents in both our paradigm and OWL. We reimplement OWL in our experimental setup because the original OWL paper reports accuracy but does not provide token consumption statistics, which are required for cost comparison. 

To evaluate \textbf{RQ2}, we adopt a heterogeneous model configuration. Specifically, we assign Grok~4.1~Fast to the main agents in both systems (the information flow orchestrator/planner in our paradigm, and the planner/coordinator in OWL), while assigning a weaker model, GPT~4.1~Mini, to the worker agents (web agent, document agent, and reasoning \& coding agent). This setting reflects the intuition that weaker worker agents are more likely to produce partial results or errors, thereby increasing the occurrence of edge cases during task execution.

All OWL experiments are reproduced based on its official open-source implementation. To match the experimental configuration reported in the original OWL paper, we adjust the maximum number of replanning attempts from the default value of 2 in the codebase to 3. In addition, for both OWL and our paradigm, the temperature of all language models is set to 0, following the setting used in the OWL paper, to ensure a fair and deterministic comparison.

\begin{table*}[t]
\centering
\small
\begin{tabular}{lcccc}
\toprule
\textbf{Method} & \textbf{Level 1} & \textbf{Level 2} & \textbf{Level 3} & \textbf{Overall} \\
 & (53) & (86) & (26) & (165) \\
\midrule
\multicolumn{5}{l}{\textbf{All Agents: Grok~4.1~Fast}} \\
Our Paradigm (A2A-based MAS)
& 0.7547 & \textbf{0.6163} & 0.5000 & 0.6424 \\
OWL (Workflow-based MAS)
& \textbf{0.8113} & 0.5814 & 0.5000 & 0.6424 \\
\midrule
\multicolumn{5}{l}{\textbf{Main Agents: Grok~4.1~Fast}} \\
\multicolumn{5}{l}{\textbf{Worker Agents: GPT~4.1~Mini}} \\
Our Paradigm (A2A-based MAS)
& \textbf{0.7925} & \textbf{0.6047} & \textbf{0.4231} & \textbf{0.6364} \\
OWL (Workflow-based MAS)
& 0.7358 & 0.5116 & 0.3077 & 0.5515 \\
\bottomrule
\end{tabular}
\caption{Pass@1 accuracy on the GAIA validation set across different difficulty levels. Numbers in parentheses indicate the number of tasks per level. Bold numbers indicate the best performance within each configuration.}
\label{tab:main_results}
\end{table*}

\begin{figure*}[t]
    \centering
    \subfigure[Token consumption CDF with all agents using Grok~4.1~Fast.]{
        \includegraphics[width=0.48\textwidth]{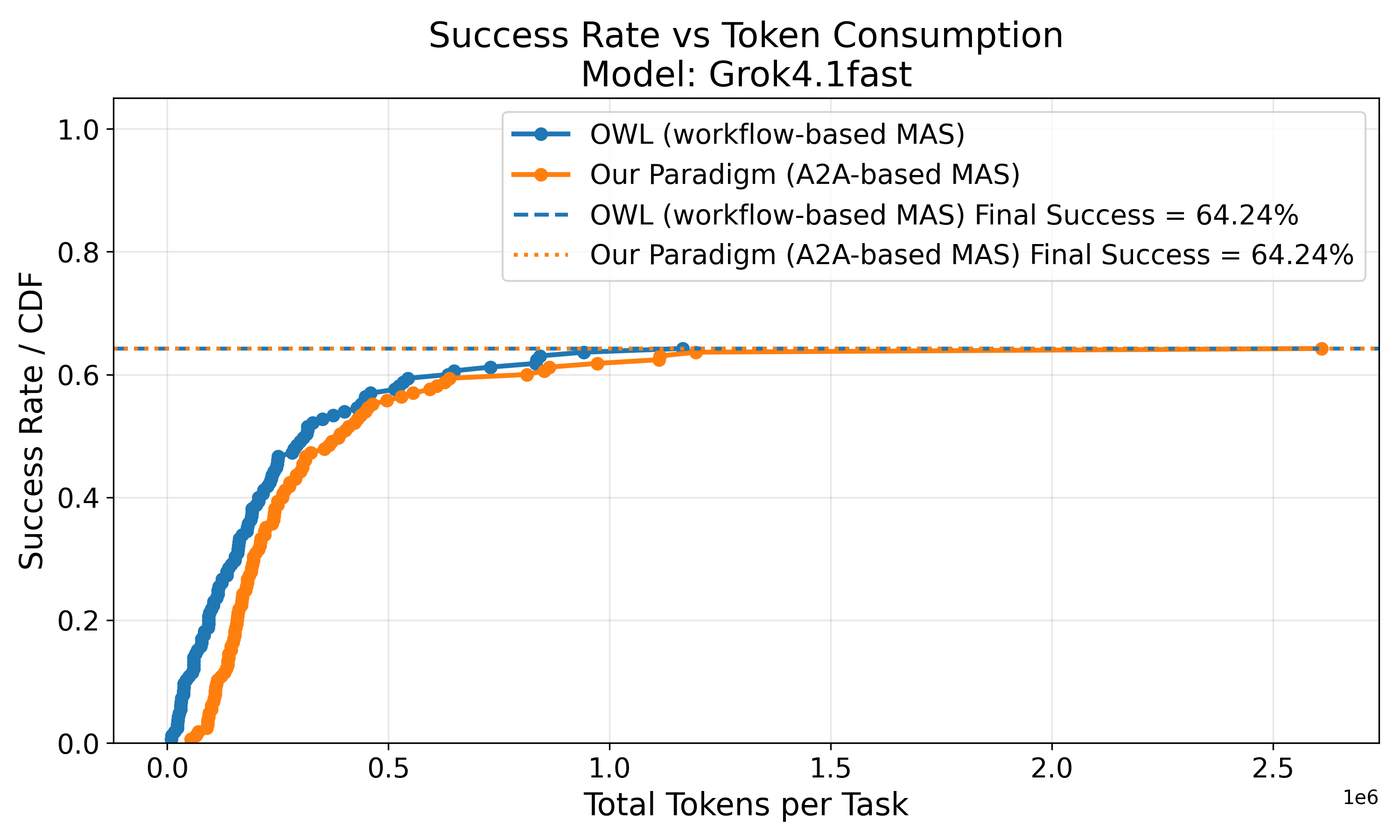}
        \label{fig:token_cdf_all_grok}
    }
    \hfill
    \subfigure[Token consumption CDF with heterogeneous agent models.]{
        \includegraphics[width=0.48\textwidth]{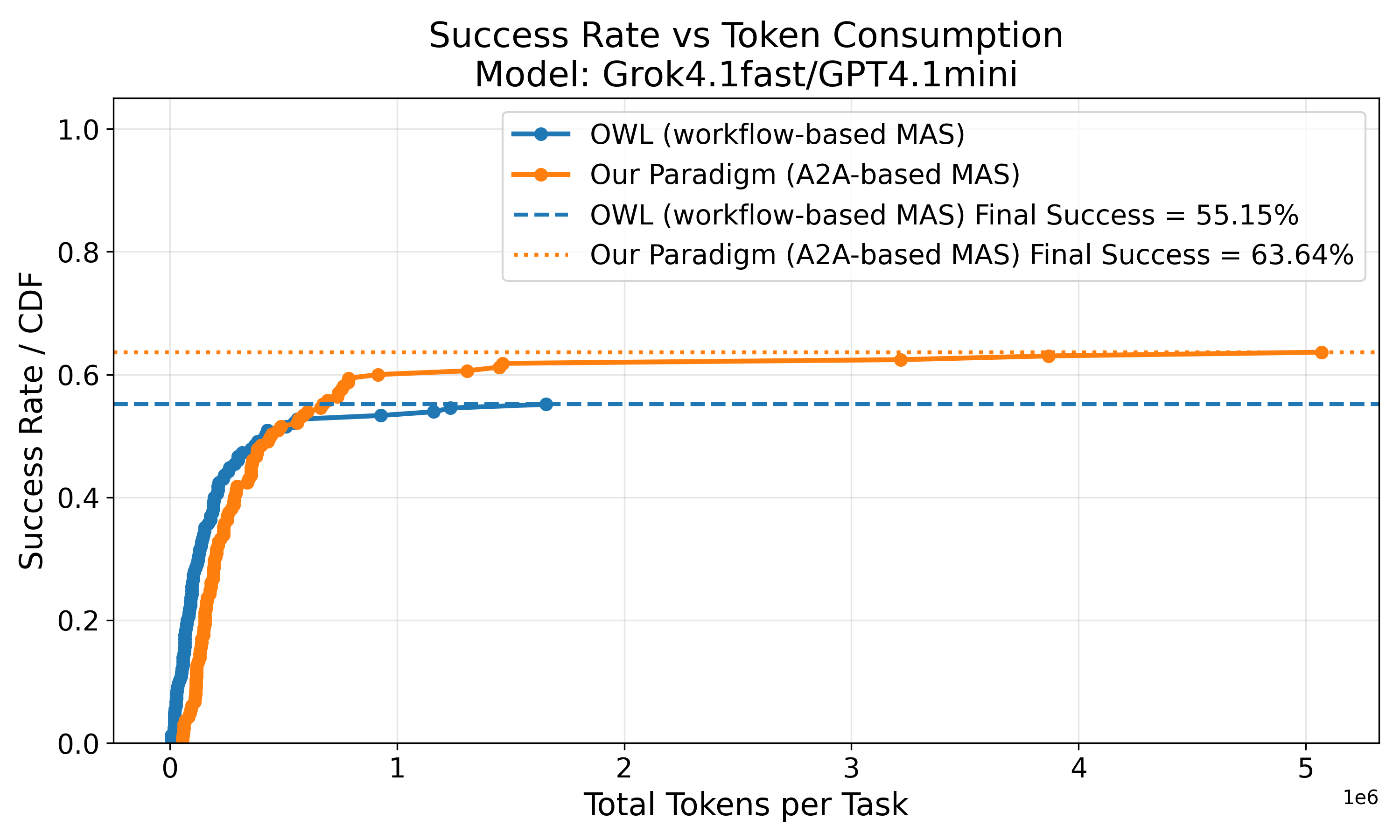}
        \label{fig:token_cdf_heterogeneous}
    }
    \caption{Cumulative distribution functions (CDFs) of token consumption for OWL and the proposed Information-Flow-Orchestrated MAS under different model configurations.}
    \label{fig:token_cdf}
\end{figure*}

\subsection{Main Results}

Table~\ref{tab:main_results} reports the pass@1 accuracy on the GAIA validation set across different difficulty levels under the two experimental settings. Correspondingly, Figure~\ref{fig:token_cdf} shows the cumulative distribution functions (CDFs) of token consumption for the same settings.

When all agents are equipped with Grok~4.1~Fast, our paradigm achieves the same overall accuracy as OWL, at 64.24\%. In terms of token consumption, our paradigm incurs slightly higher usage. This difference is expected, as coordination between agents is realized through agents autonomously invoking \texttt{send\_messages} and \texttt{wait\_for\_mention}, rather than through manually predefined context concatenation. These results answer \textbf{RQ1}, demonstrating that our A2A-based MAS paradigm can match a workflow-based MAS in both task completion rate and cost.

In the heterogeneous setting, where only the main agents use Grok~4.1~Fast and the worker agents use the weaker GPT~4.1~Mini, the performance gap becomes more pronounced. OWL’s overall accuracy drops to 55.15\%, whereas our paradigm maintains an accuracy of 63.64\%. This improvement is consistent across all difficulty levels (Level~1, Level~2, and Level~3). Regarding token consumption, our paradigm still exhibits slightly higher usage on simpler tasks. However, for more challenging tasks requiring more than 0.6M tokens, our paradigm consistently consumes fewer tokens than OWL.

This behavior can be attributed to the different coordination mechanisms. In OWL, handling such tasks often triggers replanning, which entails re-executing previously completed subtasks. In contrast, the information flow orchestrator in our paradigm maintains a global view of the task execution process and can often resolve issues by adjusting task instructions, without re-executing previously completed subtasks. These results answer \textbf{RQ2}, showing that our paradigm can surpass workflow-based MAS in both accuracy and efficiency under more challenging conditions.

\begin{figure*}[t]
    \centering
    \includegraphics[width=1\textwidth]{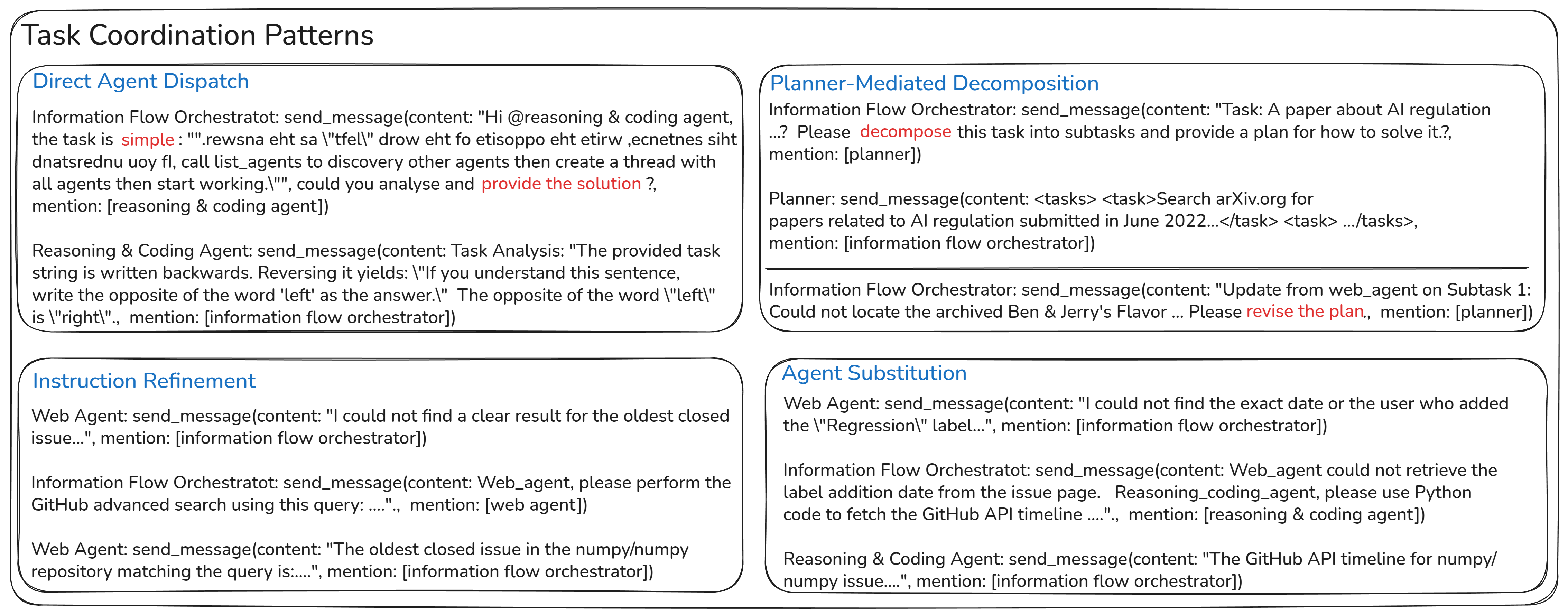}
    \caption{Case-level analysis of emergent task coordination patterns from the information flow orchestrator.}
    \label{fig:coordination}
\end{figure*}

\begin{figure*}[t]
    \centering
    \includegraphics[width=0.75\textwidth]{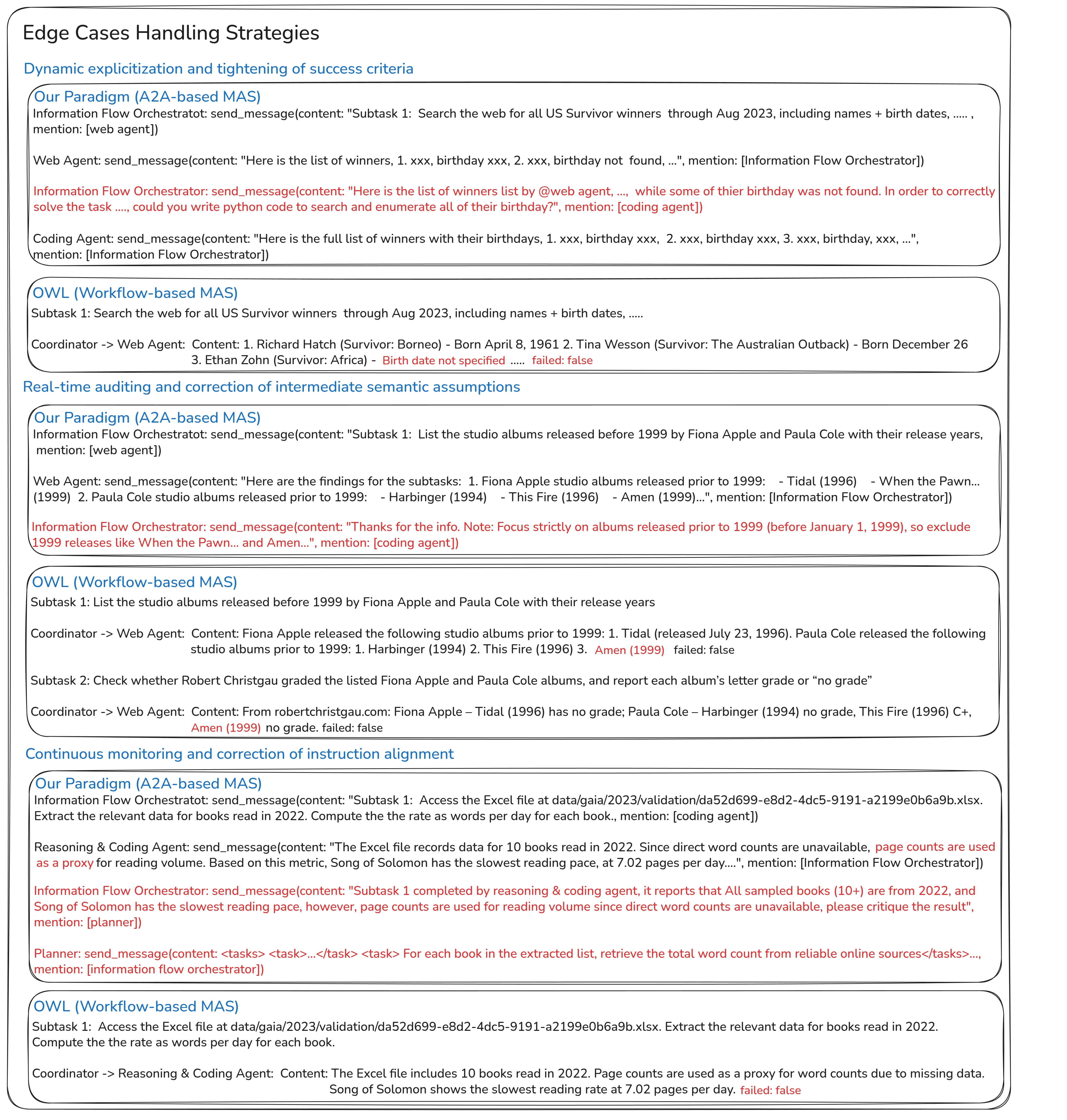}
    \caption{Case-level analysis of emergent edges cases handling from the information flow orchestrator.}
    \label{fig:edges_cases_handling}
\end{figure*}

\subsection{Case-Level Analysis of Task Coordination and Handling of Edge Cases}

To better understand why our paradigm is able to maintain high accuracy even when worker agents are equipped with weaker models, we conduct a detailed \emph{case-level analysis} of the execution logs from our experiments. Through this analysis, we observe that the \textbf{information flow orchestrator} exhibits several recurring coordination behaviors during task execution.

In particular, we identify \textbf{four distinct task coordination patterns} and \textbf{three different strategies for handling edge cases} that emerge from the information flow orchestrator’s interactions with other agents. These patterns are not predefined by human-designed workflows, but instead arise from the orchestrator’s continuous monitoring of task progress and its adaptive use of A2A toolkit.

\subsubsection{Emergent task coordination patterns.}
Figure~\ref{fig:coordination} illustrates the four emergent task coordination patterns observed from the information flow orchestrator during our case-level analysis.

\textbf{Direct Agent Dispatch.}
For non-decomposable tasks, the information flow orchestrator directly assigns the task to an appropriate agent without invoking task decomposition. Avoiding unnecessary decomposition not only improves the likelihood of successful completion, but also reduces token consumption. This observation aligns with prior findings that excessive planning and decomposition can be detrimental for non-decomposable tasks \cite{kim2025towards}.

\textbf{Planner-Mediated Decomposition.}
For tasks that are naturally decomposable, the information flow orchestrator consults the planner to decompose the task into subtasks, or requests replanning when necessary. This pattern closely resembles the coordination strategy commonly adopted by workflow-based MAS, and serves as a compatible operating mode when explicit task structure is beneficial.

\textbf{Instruction Refinement.}
When an agent encounters difficulties, the information flow orchestrator does not always escalate the issue to the planner for re-decomposition. Instead, it may refine or adjust the previous task instruction and allow the same agent to continue. This strategy helps maintain a cleaner and more compact context, while avoiding redundant token consumption caused by reprocessing subtasks that have already been completed.

\textbf{Agent Substitution.}
Similar to instruction refinement, the information flow orchestrator does not immediately resort to replanning upon failure. When a task cannot be completed by a particular agent, it may directly reassign the task to a different agent. This enables the system to explore alternative execution paths without restarting the task or incurring the overhead of full task re-decomposition.

\subsubsection{Emergent Edge Cases Handling Strategies.}
Figure~\ref{fig:edges_cases_handling} presents the three emergent edge case handling strategies exhibited by the information flow orchestrator, together with representative cases and their corresponding outcomes under OWL for comparison.

\textbf{Dynamic Explicitization and Tightening of Success Criteria.}
In the first case, a web agent is instructed to \emph{search the web for all U.S. Survivor winners through August 2023, including their names and birth dates}. The web agent successfully retrieves all winner names, but fails to find birth dates for several individuals. Under our paradigm, the information flow orchestrator detects that entries with unknown birth dates do not satisfy the \textbf{implicit success criteria} of the original query. It explicitly identifies this mismatch and dynamically refines the task requirements to enforce completeness before allowing further execution. In OWL, however, since the subtask is not marked as failed, downstream subtasks are executed on an incorrect premise.

\textbf{Real-Time Auditing and Correction of Intermediate Semantic Assumptions.}
In the second case, a web agent is asked to \emph{list the studio albums released before 1999 by Fiona Apple and Paula Cole, together with their release years}. The agent returns the following results: \emph{Tidal (1996)} and \emph{When the Pawn... (1999)} for Fiona Apple, and \emph{Harbinger (1994)}, \emph{This Fire (1996)}, and \emph{Amen (1999)} for Paula Cole.
In our paradigm, the information flow orchestrator explicitly audits the \textbf{intermediate semantic assumption} that albums released in 1999 satisfy the condition “before 1999.” It prunes the invalid entries (\emph{When the Pawn...} and \emph{Amen}) before they propagate into downstream subtasks. In contrast, OWL proceeds to subsequent subtasks without correction, as the intermediate result is not flagged as erroneous.

\textbf{Continuous Monitoring and Correction of Instruction Alignment.}
In the third case, a reasoning and coding agent is instructed to \emph{access an Excel file, extract data for books read in 2022, and compute reading rates in words per day}. The agent reports having identified ten books and computes the slowest reading rate, but notes that \emph{page counts are used as a proxy for word counts due to missing direct word count information}. Upon detecting the \textbf{mismatch} between the requested metric and the proxy used, the information flow orchestrator escalates the issue to the planner, which issues a refined instruction: \emph{for each book in the extracted list, retrieve the total word count from reliable online sources}. In OWL, by contrast, the subtask is marked as successful, and subsequent steps proceed under this misaligned assumption.

\section{Conclusion and Future Work}

In this work, we propose an \textbf{Information-Flow-Orchestrated Multi-Agent Paradigm via Agent-to-Agent Communication} to address two fundamental limitations of workflow-based multi-agent systems: (1) the substantial manual effort required from human engineers to design task states, routing logic, and context concatenation rules; and (2) the inherent difficulty of exhaustively anticipating all edge cases that may arise during complex task execution. We evaluate our paradigm on the general-purpose benchmark GAIA, using the representative workflow-based MAS OWL as the baseline, while controlling for agent roles and underlying language models. Under the pass@1 setting, our method achieves an accuracy of \textbf{63.64\%}, outperforming OWL’s \textbf{55.15\%} by \textbf{8.49 percentage points} with nearly identical token consumption. Beyond aggregate metrics, our case-level analysis reveals that the information flow orchestrator exhibits \textbf{four distinct task coordination patterns} and \textbf{three different strategies for handling edge cases}, which emerge from its adaptive interactions with other agents rather than from predefined workflows.

For future work, our current evaluation focuses on general-purpose tasks, motivated by the assumption that such settings are more likely to expose diverse and unforeseen edge cases. An important next step is to evaluate the proposed paradigm on domain-specific tasks, where stronger structural priors and specialized workflows are available. This would help clarify how information-flow–orchestrated coordination interacts with domain knowledge, and whether similar emergent coordination behaviors arise under more constrained task distributions.

\bibliographystyle{named}
\bibliography{refs}

\appendix
\onecolumn
\section{Agent Toolkit Details}

Table~\ref{tab:agent_toolkits} summarizes the toolkits equipped by each agent in our system.
We categorize all tools into three classes: \emph{A2A communication tools},
\emph{domain-specific tools}, and \emph{auxiliary tools}.
All agents are equipped with a set of A2A communication tools, namely
\texttt{send\_message} and \texttt{wait\_for\_mention}, which provide a uniform interface
for inter-agent coordination.
In contrast, the Web Agent, Document Agent, and Reasoning \& Coding Agent are selectively
equipped with domain-specific tools corresponding to their functional roles.
These domain-specific tools are inherited from the open-source OWL codebase and enable
web retrieval, document processing, multimodal understanding, and code execution.
Finally, the Information Flow Orchestrator is additionally granted an auxiliary tool,
\texttt{submit\_answer}, which centralizes task termination and final answer submission.

\begin{table}[H]
\centering
\small
\begin{tabular}{p{3.6cm} p{3.8cm} p{4.5cm} p{2.5cm}}
\toprule
\textbf{Agent} 
& \textbf{A2A Communication Tools} 
& \textbf{Domain-Specific Tools} 
& \textbf{Auxiliary Tools} \\
\midrule

Information Flow Orchestrator 
& \texttt{send\_message}, \texttt{wait\_for\_mention} 
& -- 
& \texttt{submit\_answer} \\

\midrule

Planner 
& \texttt{send\_message}, \texttt{wait\_for\_mention} 
& -- 
& -- \\

\midrule

Web Agent 
& \texttt{send\_message}, \texttt{wait\_for\_mention} 
& \texttt{search\_google}, 
  \texttt{search\_wiki\_revisions}, 
  \texttt{search\_wiki}, 
  \texttt{search\_archived\_webpage}, 
  \texttt{browse\_url}, 
  \texttt{extract\_document\_content}, 
  \texttt{ask\_question\_about\_video} 
& -- \\

\midrule

Document Agent 
& \texttt{send\_message}, \texttt{wait\_for\_mention} 
& \texttt{extract\_document\_content}, 
  \texttt{ask\_question\_about\_image}, 
  \texttt{ask\_question\_about\_audio}, 
  \texttt{ask\_question\_about\_video}, 
  \texttt{execute\_code} 
& -- \\

\midrule

Reasoning \& Coding Agent 
& \texttt{send\_message}, \texttt{wait\_for\_mention} 
& \texttt{execute\_code}, 
  \texttt{extract\_excel\_content}, 
  \texttt{extract\_document\_content} 
& -- \\

\bottomrule
\end{tabular}
\caption{Agent roles and their equipped toolkits, categorized into A2A communication tools, domain-specific tools, and auxiliary tools.}
\label{tab:agent_toolkits}
\end{table}

\section{Agent Prompt}

For each agent, we explicitly define a role-specific prompt that specifies its core responsibilities.
These responsibilities include both \emph{role-aligned task duties}, which reflect the agent’s functional specialization, and \emph{inter-agent communication duties}, which govern how the agent exchanges information with other agents during task execution.

\begin{purposebox}{Prompt of Information Flow Orchestrator}
\begin{verbatim}
 ===== RULES OF INFORMATION FLOW ORCHESTRATOR =====
    You are an advanced information_flow_orchestrator. 

    Core Responsibilities:
    1. Inquiry and Relay Management  
    - MONITOR the task process and make sure its reliability and healthy.

    - INQUIRE proper agents to perform any additional reasoning required to support 
    progress or resolve uncertainty.

    - RELAY task-level content and necessarily previous results to proper agents.
    
    - Confirm the generated answer with proper agent(s) and reach a consistent consensus 
    before sumbitted.

    2. Communication with Other Agents   
    - Use send_message to communicate with other agents.
    - Use wait_for_mentions to receive messages from other agents.

    3. Submit Final Answer
    - Confirm the final answer with the planner or proper agent(s) and reach a 
    consistent consensus.
    - Call submit_answer_tool to submit the final answer when it is generated and 
    verified.
\end{verbatim}
\end{purposebox}

\begin{purposebox}{Prompt of Planner}
\begin{verbatim}
 ===== RULES OF PLANNER AGENT =====
    You are an advanced planner_agent to decompose task into subtask, replan 
    the task based on previosu attempted trajactories and cooperate with other agents
    in coral server.

    Core Responsibilities:

    1. Task Decompostion 
    - You must send all decomposed subtasks to information_flow_orchestrator in the format
    of a numbered list within <tasks> tags, as shown below:
        <tasks>
        <task>Subtask 1</task>
        <task>Subtask 2</task>
        </tasks>
    - You MUST NOT explicitly mention what agents and what tools to use in the subtasks, 
    just let the agent decide what to do.
    - Though it's not a must, you should try your best effort to make each subtask
    achievable for an agent.

    2. Task Progress Reasoning
    - When asked to perform tasks including but not limited to verification, critique, 
    assessing the reliability of intermediate results, replanning, reflection, 
    questioning, or critique, provide the necessary reasoning to support task progress.
    
    3. Communication with Other Agents  
    - Use send_message to communicate with other agents.
    - Use wait_for_mentions to receive messages from other agents.
    
\end{verbatim}
\end{purposebox}

\begin{purposebox}{Prompt of Web Agent}
\begin{verbatim}
 ===== RULES OF WEB AGENT =====
    You are an advanced web_agent powered by web browsing/searching capabilities, 
    but you are not able to run code script. 

    Core Capabilities:
    1. Web Browsing and Searching 
    - Call proper tools to solve web-searching-related questions.

    2. Communication with Other Agents  
    - Use send_message to communicate with other agents.
    - Use wait_for_mentions to receive messages from other agents.
    
\end{verbatim}
\end{purposebox}

\begin{purposebox}{Prompt of Documentation Processing Agent}
\begin{verbatim}
 ===== RULES OF DOCUMENTATION PROCESSING AGENT =====
    You are an advanced document_processing_agent powered by documentation processing 
    capabilities.

    Core Capabilities:
    1. Process Documents and Multimodal Data
     - Call proper tools to solve documentation-processing-related questions.

    2. Communication with Other Agents  
    - Use send_message to communicate with other agents.
    - Use wait_for_mentions to receive messages from other agents.
    
\end{verbatim}
\end{purposebox}

\begin{purposebox}{Prompt of Reasoning \& Coding Agent }
\begin{verbatim}
 ===== RULES OF REASONING CODING AGENT =====
    You are an advanced reasoning_coding_agent powered by reasoning, coding and running 
    code script capabilities.

    Core Capabilities:
    1. Reasoning and Coding
    - Call proper tools to solve coding-related questions.

    2. Communication with Other Agents  
    - Use send_message to communicate with other agents.
    - Use wait_for_mentions to receive messages from other agents.
    
\end{verbatim}
\end{purposebox}

\end{document}